\documentclass[conference]{IEEEtran}
\usepackage{times}

\usepackage[numbers]{natbib}
\usepackage{multicol}
\usepackage[bookmarks=true]{hyperref}
\usepackage{graphicx}
\usepackage{amsmath}
\usepackage[position=bottom,labelfont=bf,textfont={sl,bf},lofdepth,lotdepth]{subfig}
\usepackage{amssymb}
\usepackage{gensymb}
\usepackage{tikz}
\usepackage{pgf}
\usetikzlibrary{fit,positioning,calc,arrows}
\usepackage[]{algorithm2e}
\usepackage{color,soul}

\begin{document}

\title{Spatiotemporal Articulated Models for \\ Dynamic SLAM}

% You will get a Paper-ID when submitting a pdf file to the conference system
\author{\authorblockN{Suren Kumar, Vikas Dhiman, Madan Ravi Ganesh, Jason J. Corso}
	\authorblockA{Electrical Engineering and Computer Science,\\
	University of Michigan, Ann Arbor, MI - 48109\\
$\{$surenkum, dhiman, madantrg, jjcorso$\}$@umich.edu}
}

%\author{\authorblockN{Michael Shell}
%\authorblockA{School of Electrical and\\Computer Engineering\\
%Georgia Institute of Technology\\
%Atlanta, Georgia 30332--0250\\
%Email: mshell@ece.gatech.edu}
%\and
%\authorblockA{Twentieth Century Fox\\
%\and
%\authorblockN{James Kirk\\ and Montgomery Scott}
%\authorblockA{Starfleet Academy\\
%San Francisco, California 96678-2391\\

% avoiding spaces at the end of the author lines is not a problem with
% conference papers because we don't use \thanks or \IEEEmembership

%Homer Simpson\authorrefmark{2},
%Montgomery Scott\authorrefmark{3} and
%Eldon Tyrell\authorrefmark{4}}
%Telephone: (800) 555--1212, Fax: (888) 555--1212}
%\authorblockA{\authorrefmark{4}Tyrell Inc., 123 Replicant Street, Los Angeles, California 90210--4321}}

\maketitle
\IEEEoverridecommandlockouts
\begin{abstract}
	We propose an online spatiotemporal articulation model estimation
	framework that estimates both articulated structure as well as a 
	temporal prediction model solely using passive observations. 
	The resulting model can predict future
	motions of an articulated object with high confidence because of the spatial
	and temporal structure. We demonstrate 
	the effectiveness of the predictive model by incorporating it within 
	a standard simultaneous localization and mapping (SLAM) pipeline for 
	mapping and robot localization in previously unexplored dynamic environments. 
	Our method is able to localize the robot and map a dynamic scene by 
	explaining the observed motion in the world.
	We demonstrate the effectiveness of the proposed framework for both
	simulated and real-world dynamic environments.
%Articulated motion understanding is one of the fundamental challenges which impacts
%a wide-variety of robotics research including robot navigation, object 
%tracking and robot-environment interaction. General rigid body motion has $6$
%degrees of freedom which is constrained by the articulated structure, such as,
%motion of a door constrained by a hinge. The representation of such constraints
%is employed to generate accurate prediction of rigid bodies which enables 
%navigation in dynamic environments and object tracking \cite{schmidt2014dart}.
%Constraints induced via articulated structure are also important to guide
%robotic manipulation task, such as, opening a drawer \cite{burget2013whole}. 
%Despite the overwhelming importance of articulated structure and motion
%understanding, there has been limited work to address this problem for unknown
%environments without any apriori knowledge. To that end, we propose spatiotemporal 
%articulated simultaneous localization and mapping \textit{STA-SLAM} which 
%explicitly represents the articulation present in the environment while
%simultaneously localizing the robot in an unknown environment.
\end{abstract}

\IEEEpeerreviewmaketitle

\section{Introduction}

Simultaneous localization and mapping (SLAM) has significantly improved over the 
last few decades, moving from sparse, $2D$, slow feature-based maps 
\cite{smith1986representation} to dense, $3D$, fast maps 
\cite{newcombe2011kinectfusion}.
For example, KinectFusion \cite{newcombe2011kinectfusion} impressively estimates 
a full $6$ degrees-of-freedom (DOF) camera motion while building a dense, 
metrically accurate $3D$ map of an environment in real-time 
\cite{newcombe2011kinectfusion}. However, most of these compelling developments 
in SLAM operate under the assumption that the environment is static; i.e., any 
motion in the environment is considered to be noise. Yet, most environments in 
our world are dynamic---roadways, hospitals, factories, etc. The ability for 
robotic systems to deal with such motion is increasingly critical for the 
advancement of robotic systems used in autonomous driving \cite{levinson2011towards}, 
interaction with articulated objects \cite{burget2013whole} and tracking 
\cite{schmidt2014dart}, among others.

Early attempts at extending SLAM to dynamic environments either explicitly track 
the moving objects \cite{wang2003online}, remove the moving landmarks 
\cite{hahnel2003mobile} from observations, or, alternatively, build two 
different maps for static and dynamic parts of the environment as Wolf et al.  
\cite{wolf2005mobile} do. However, all of these methods rely strictly on the 
static parts of the environment to localize the robot. As a 
consequence, these approaches would fail to extend to environments with 
significant motion. Fundamentally, neglecting the dynamic part of the 
environment for robot localization implies that observing the motion provides no 
information about the location of the robot.

On the contrary, we observe that motion has a rich structure that can be 
leveraged for improving SLAM. We will use the running example of a typical, 
hinged door throughout this paper. The motion of this door has just a single 
DOF---the angle about the hinge---rather than the full 6 DOF. 
Intuitively, observing the relative motion of a door constrains the possible 
locations of a robot, even though the door itself may be moving. More 
concretely, explaining away (removing) the motion of a door by explicitly 
modeling it provides the same constraint for the localization problem as 
observing a static landmark. 

To that end, we propose a two-part method for joint articulated motion 
estimation and SLAM in dynamic environments. Our proposed method first estimates 
the articulated structure, such as, axis of the motion of a door and its motion 
parameters, such as, current angle of the door; and, it does so spatiotemporally
in an online manner.
Essentially, the first step finds the best subspace of the special euclidean 
manifold ($SE(3)$) to represent the motion of an object. For our door example, 
which is a $1$ DOF revolute joint, the subspace is a unit circle.  All such 
observed motion in the scene is eventually associated with motion subspace, and 
each such subspace further constrains the SLAM problem.  This first part of our 
method is inspired by psychophysical experiments on human motion understanding 
that have demonstrated humans first distinguish between competing articulations 
(translation, rotation, and expansion) and then estimate the motion conditioned 
on the motion model~\cite{NIPS2008_3458}.

In the second part, we build motion-continuity-based temporal prediction models 
to capture the temporal evolution of the motion
parameters in the established subspace. 
To summarize, the first part finds a $1$ DOF subspace of the 
special euclidean ($SE(3)$) manifold and the second part models the evolution of 
an object's motion in the resulting subspace. The resulting framework allows us 
to estimate articulated structure along with the motion propagation model and 
hence, allows us to include articulated motions in an online SLAM framework.
%
% I think the following sentence needs to be expanded, but I struggle to find 
% the exact words
Indeed, we show the application of temporal predictions of the resulting model 
for SLAM in dynamic environments. 

To the best of our knowledge, this is the first method that explicitly estimates 
spatiotemporal articulated motion and integrates it while performing SLAM in a 
dynamic environment. The proposed method has two major contributions. First, we 
propose an online articulated structure with an explicit temporal model 
estimation from a low number of feature tracks.  Theoretically, our method can estimate 
articulated structure of a prismatic or revolute articulated joint using just a 
single feature track. Second, we show how the spatiotemporal articulated motion 
prediction can be integrated within SLAM to estimate the motion of the 
dynamic objects while building a map of an unknown dynamic environment. We 
evaluate the proposed framework with regards to articulation structure, temporal 
predictions, and SLAM capabilities in both simulation and real-life experiments.

\section{Related Work}
We review previous work related to both spatiotemporal articulated structure estimation as
well as SLAM in dynamic environments.

\subsection{Articulation Structure} 
Articulation structure estimation methods attempt to recover the linking
structure, or kinematic chain, of rigid bodies, essentially discovering the articulated joints that
constrain the motion of rigid bodies to a subspace of $SE(3)$.
The early approaches to address this problem extended multibody structure
from motion \cite{costeira1998multibody} ideas to understand articulation by
clustering feature trajectories into individual rigid bodies
\cite{yan2006automatic}. With the introduction of the commercial depth camera, the feature
trajectories could be directly represented in $3D$ and thus avoiding the need to
estimate shape \cite{Pillai-RSS-14,katz2013interactive}. However, these methods
implicitly assume that large number of features trajectories are available 
whereas our model can estimate the articulation structure of a single trajectory.

To avoid the problem of dense feature tracking, some methods achieve direct 
sensing of articulated motion via placement of markers, such as, ARToolKit 
\cite{fiala2005comparing}, checker-board markers etc. 
\cite{gray2013single,sturm2011probabilistic,sturm2013learning,hausman2015active}.  
The placement of markers removes the need for an otherwise noisy feature-tracking 
process from the structure estimation problem.  Another way to get better 
estimation of articulated motion is via active interaction of a robot 
manipulating an articulated object \cite{katz2008extracting, hausman2015active}.  
In contrast, our method requires no such markers nor active manipulation; in 
fact, using our method, a robotic system could observe a different agent 
manipulating the environment to infer its articulation structure.

These prior works in estimating articulation structure have mostly relied on 
collecting data from demonstrations and performing articulation estimation 
offline, e.g., \cite{sturm2011probabilistic,sturm2013learning}. 
In contrast to 
these state-of-the-art methods, we do online articulation estimation.  Online 
estimation not only enables evolving beliefs with more observations but also 
allows for inclusion in online tracking and mapping algorithms.  The closest 
work to ours is Martin et al.  \cite{martin2014online} who propose a framework 
for online estimation; however, their method has no explicit probabilistic 
measure for model confidence to select an articulation model. Furthermore, the 
state-of-the-art in articulation estimation does not model the
temporal evolution of motion.

Our model directly addresses this lack of temporal
modeling (for example, acceleration/deceleration of a door) in articulation 
estimation. We
propose an explicit temporal model for each articulation type, which is 
necessary
to make good long-term future predictions. Temporal modeling of arbitrary order
allows us to: i) track new parts/objects that enter/exit the scene
\cite{martin2014online};  ii) model the entire scene and as a result explore
dependencies between neighboring objects; and, iii) assimilate articulated
object motion in SLAM
\cite{durrant2006simultaneous}. Apart from the applications presented in this 
paper, temporal models associated with articulated structure will help in 
robot-environment interaction, specifically obtaining dynamic characteristics of 
the objects in the environment \cite{sturm2011probabilistic}.

\subsection{SLAM in Dynamic Environments} 

Previous literature to handle dynamic environments can be divided into two 
predominant approaches: i) detect moving objects and ignore them \cite{hahnel2003mobile}, and ii) track moving objects as landmarks 
\cite{bailey2006simultaneous}. In the first approach, using the fact that the 
conventional SLAM map is highly redundant, the moving landmarks can be removed 
from the map building process \cite{bailey2002mobile}. In contrast, 
moving-object tracking-based approaches explicitly track moving objects by 
adding them to the estimation state~\cite{wang2003online}. Hahnel et al. 
\cite{hahnel2003mobile} propose tracking humans in dense populated environments 
and removing the areas corresponding to humans to build static maps. Nieuwenhuisen 
et al. \cite{nieuwenhuisen2010improving} proposed using a parametrized model of 
a door to localize doors in a moving environment while simultaneously 
estimating a static map.  Recent work has focused on updating the map by 
removing dynamic parts of the scene without considering the nature of motion in 
the environment \cite{ferri2015dynamic}. However, these methods assume that the 
measurement of moving objects carries no useful information for the robot 
localization.  Furthermore, these method do not strive to build maps of dynamic 
environments that can be used to discover articulated structure or novel 
objects.  Our method demonstrates the utility of estimating motion and 
incorporating it into SLAM in both simulated and real-world cases.

Stachniss and Burgard \cite{stachniss2005mobile} considered a graphic model similar to ours to update the
map of a dynamic environment by using local patch maps and modeling transitions
between patch maps. However, such an approach is only
suitable for quasi-static environments as the number of maps required would
increase exponentially with the number and pose characterizations of dynamic objects in the
world. 
A recent work has extended dense tracking and mapping to
dynamic scenes by estimating a dense warp field
\cite{newcombe2015dynamicfusion}. However, the approach disregards the object
level rigid nature which limits its application to topological changes, such as,
open to closed door.
In the context of manipulating doors, Pretrovskaya et al.
\cite{petrovskaya2007probabilistic} used a model of the door and a prior
low-resolution static map of environment to track objects for manipulation tasks.
In contrast, our approach does not need any prior maps of the environment or
articulated objects, and does not suffer from exponential increase in the number of maps for a dynamic
environment. 

\section{Spatiotemporal Articulation Models} 
\subsection{Articulation Models}
We represent all articulated motion in the
environment as 
\begin{align}
z_t = f_M(C,q(t)) + \nu \label{eq:articulated_model_representation}
\end{align}
where $z_t$ is the observed motion of an object (a trajectory in $3D$ at time
$t$), $M \in \{M_j\}_{j=1}^r$ is one of the $r$ possible motion models, $C$ is
the configuration space (e.g. axis vector in case of a door), $q(t)$ represents
the time-varying motion variables (e.g. length of prismatic joint, angle of
door) associated with the motion model $M$ and $\nu$ is the noise in
observations. We assume that all observed motion can be explained by one 
of the pre-defined $r$ possible motion models. Indeed, there are only finite
articulated joint models in man-made environments.

This kind of representation, where non-time varying configuration parameters
are separated from time-varying motion variables, is beneficial for a variety of
reasons: First, it allows for the unified treatment of various types of articulation
due to its single and consistent representation of motion variables. Second,
this representation can be robustly estimated from experimental data given the
reduced number of parameters to be estimated. Furthermore, it often makes the
estimation problem linear, and as a consequence, convex.

A notable omission from our modeling of articulation model, Eq.
\ref{eq:articulated_model_representation} is the input to the system such as
torque acting on a door, force on a drawer etc. This modeling limitation is due
to the passive nature of our sensing approach in addition to no prior
information about the agents in the scene. To probabilistically predict motion
at the next time step $P(z_{t+1}|z_t)$ (under Markovian assumption) without modeling the input forces/torques
(thus not using a dynamics model), we model the propagation of motion
variables $P(q_{t+1}|q_t)$. 

\begin{figure}
\includegraphics[width=1\linewidth,trim = 0mm 85mm 40mm
0mm,clip]{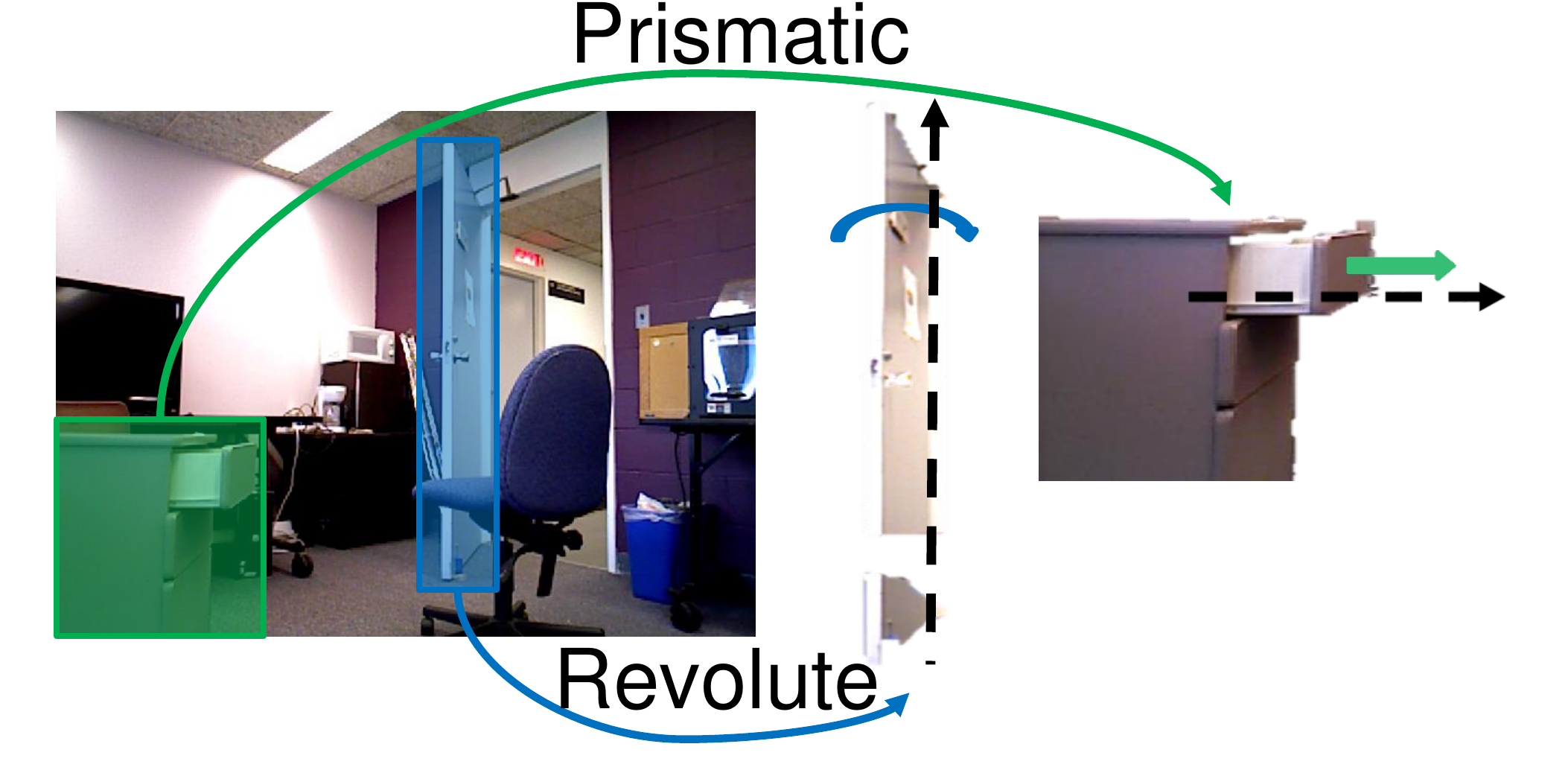}
\caption{Demonstration of articulated joints considered in this work as seen in
a typical indoor environment. Revolute and prismatic joints are 1 DOF joints,
whose articulated structure is the axis of motion and the motion parameter is the
angle and length along the axes respectively. }
\label{fig:articulation_classification}
\end{figure}

The configuration parameters in Eq. \ref{eq:articulated_model_representation}
are entirely dependent on the type of articulated joint.  In this section, we
consider the problem of articulation identification from point correspondences
over time. Rigid bodies can move in $3D$ space with $SE(3)$ configuration, which
is the product space of $SO(3)$ (Rotation Group for 3D rotation) and $E(3)$
(Translation using 3D movement). The full $SE(3)$ has $6$ DOF, which are reduced 
when a rigid body is connected to another rigid body
via a joint. For example, the configuration space for a revolute joint (1 DOF
joint) can be assumed to be a connected subset of the unit circle. Figure
\ref{fig:articulation_classification} shows some of the articulated joints modeled in this work.

We model the environment as collection of landmarks which can move according to
three different articulation models: static, prismatic and revolute.
To find a relevant subspace of $SE(3)$, we fit a circle and line in $3D$ using
least square formulation (optimal for Gaussian noise) to
estimate the subspace corresponding to revolute and prismatic joint
respectively. For the static joint, we model a perturbation from the previous
position as motion variable. We provide more details on the articulated models
used in the current
paper in Appendix \ref{sec:jac_models}.

\subsection{Temporal Structure}\label{sec:temporal_structure} Articulation
estimation provides us with configuration parameters of the articulated motion
but one still needs to estimate the evolution of motion variables over time
(e.g. position of the object along an axis for a prismatic joint).
%Temporal propagation of articulated bodies will require knowledge of dynamic model parameters (mass, friction etc.) apart from the external excitation (motor torque, force) applied  to the system. Several approaches have been proposed for estimating these parameters that use ground truth trajectories to estimate inertial and friction parameters \cite{endres2013learning}, but they assume a priori access to the object. Furthermore, the external excitation cannot be predicted as it might vary depending on the intention of agents. 
The goal of our approach is to enforce a structure on the evolution of articulated motion without using any prior information specific to the current articulated body. We take our inspiration from neuroscience literature which posits that humans produce smooth trajectories to plan movements from one point to another in environment \cite{flash1985coordination}. This smoothness assumption can be leveraged by using motion models that use limited number of position derivatives. 

Let us assume that $q(t)$ is the articulated motion variable (e.g. extension of
a prismatic joint, angle of door along a hinge ). The system model for a finite
order motion in discrete time domain with $\mathbb{X}(t) = [q,q^{(1)},.,.,q^{(n-1)}]$ (dropping the explicit time dependence of $q$ and using superscript to denote the order of derivative) as the state can be written as,
%\begin{align}
%\begin{bmatrix}
%    q^{(1)} \\
%    q^{(2)} \\
%. \\
%.\\
%q^{(n)}
%\end{bmatrix} = 
%\begin{bmatrix}
%0 & 1  & . & . & 0\\
%0 & 0  & . & . & 0\\
%0 & 0  & . & . & 0\\
%0& 0 & . & . & 1\\
%0 & 0  & . & . & 0\\
%\end{bmatrix}
%\begin{bmatrix}
%q \\
%q^{1} \\
%. \\
%.\\
%q^{n-1}
%\end{bmatrix}
%+\begin{bmatrix}
%0 \\
%0 \\
%.\\
%.\\
%1
%\end{bmatrix} \eta
%\end{align}
%This state propagation model can be converted to a discrete time model as 
\begin{align}
	&\mathbb{X}(t+1) = A \mathbb{X}(t) + B \eta \nonumber\\
	       &A = 
\begin{bmatrix}
 1 & \delta  t & \frac{{\delta  t}^2}{2} &  . & . \\
0 & 1 & \delta  t &  . & . \\
0 & 0 & 1  &  . & . \\
0 & 0 & 1  &  . & . \\
0 & 0 & .  &  . & . \\
0&  0&    0 &     0 &    1
\end{bmatrix}
B = 
\begin{bmatrix}
\frac{{\delta t}^n}{n!}\\
\frac{{\delta t}^{n-1}}{(n-1)!}\\
.\\
.\\
\frac{{\delta t}^2}{2!}\\
\delta t
\end{bmatrix}\label{eq:motion_parameter_prop}
\end{align}
%From : http://www.lehigh.edu/~eus204/Teaching/ME433/lectures/lecture07_handout.pdf

where $q^{(n)}$ denotes $n^{th}$ order derivative of the motion variable and $\eta$ is the noise. 
%where $A$ is simply the matrix exponential $\exp\{A^c\delta t\}$ of the matrix representation $A^c$  in continuous time and $B = (\int_0^{\delta T}\exp\{A^c\mu\}\mathrm{d}\mu) B^c $ with $B^c$ being the corresponding continuous time representation. 

The model considered in Eq.~\ref{eq:motion_parameter_prop} attempts
to predict the motion variable $q(t+1)$ using the information at time step $t$. 
It is a finite order Taylor series expansion of the motion variable $q(t+1)$. 
%\begin{align}
%    q(t+\delta t) = q(t) + \frac{q^{(1)}}{1!}\delta t + \frac{q^{(2)}}{2!}(\delta t)^2 + . . . +\sum_{k=n}^{\infty} \frac{q^{(k)}}{n!}(\delta t)^k \label{eq:taylor_approx}
%\end{align} 
We initialize the state $\mathbb{X}$ by obtaining $q(t)$ via inverse
kinematics based on Eq. \ref{eq:articulated_model_representation} and then
evaluating numerical derivatives via Gaussian convolved motion parameter estimates.
Our representation assumes the differentiability of the motion variable. The approximation error is of the order $O((\delta t)^n)$. Various convergence studies can be done to choose the right order $n$ for a given time duration $\delta t$, but here we study the physical aspects of the problem.

\paragraph{Choosing Order}
Ideally, one would want to choose the motion variable order as high as possible
to reduce the approximation error in Taylor series, especially, for long-term
behavior prediction which is necessary for motion planning or when sensors go
blind. But, higher order motion models will result in over-fitting because of the need to estimate more parameters from few initial samples. It also increases the filtering problem 
complexity as Kalman filtering involves matrix multiplication. Hence, the 
computational complexity is atleast $O(N^2)$, where $N$ is the length of 
the state vector. Furthermore, the error in estimating higher-order 
derivatives of a noisy signal increases exponentially with respect to derivative order.

For the current work, we found the first order motion models to be adequate for
most of the experimental use cases.
%However, there is a number of reasons why we might get away with choosing a smaller order of temporal variables. First, in classical mechanics, we only consider second order derivatives of position variables. Also humans minimize jerk \cite{flash1985coordination} in their motion.

%Following the  \cite{flash1985coordination}, we assume jerk to be the noise in the system models. Using this method of providing temporal structure to the motion, we can write the motion of a prismatic joint with state $\mathbb{X}(k) = [x[k],y[k],v[k],\dot{v}[k],\ddot{v}[k],\dddot{v}[k]]$ as 
%\begin{align}
%\mathbb{X}(k+1) = 
%\begin{bmatrix}
%1 & 0 & \cos{\theta} & 0 & 0\\
%0 & 1 & \sin{\theta} & 0 & 0\\
%0 & 0 & 1 & \delta t &\frac{{\delta  t}^2}{2}\\
%0 & 0 & 0 & 1 &\delta t \\
%0 & 0 & 0 & 0 &1 \\
%\end{bmatrix} \mathbb{X}(k)+
%\begin{bmatrix}
%0\\
%0\\
%0\\
%0\\
%1
%\end{bmatrix}\eta
%\end{align}
%where $\theta$ is the direction of prismatic axis in 2D. The covariance of noise $\eta$ in jerk needs to be continuously updated \cite{castella1980adaptive} (maybe even least squares with forgetting?) to enable one to track all the possible range of smooth motions that can be performed by an articulated object.

\section{Articulation Model Estimation}
We now consider the task of estimating the type of articulated model $M \in
\{M_j\}_{j=1}^r$ out of $r$ different models. This does not automatically follow
from the configuration and motion variables estimation. For example, consider
the case of a point particle moving in $3D$ space: one can fit a line, circle or
assume it to be static. One can potentially use goodness-of-fit measures to
estimate the appropriate model along with some heuristics. However, there are
various limitations in comparing goodness-of-fit measures related to the number
of free parameters in different models, noise in the data, over-fitting and the
number of data samples required \cite{schunn2005evaluating}. Instead of picking
a model at the initial time-step, we use a filtering-based multiple model approach to correctly pick the model for a given object.

We assume that our target object/particle obeys one of the $r$ 
different motion models. In the current formulation, we assume a uniform
prior $\mu_j(0) = P(M_j), \sum_{j=1}^{r}\mu_j(0) = 1$, over different
spatiotemporal articulation models for each individual object. This prior can be
modified appropriately by object detection. For example, a door is more likely
to have a revolute joint. Motion model probabilities are updated as more and
more observations are received using laws of total probability
\cite{yaakov2001estimation}. 
\begin{align}
& \mu_j(t) \equiv P(M_j|\mathbf{Z}_{0:t})  = 
\frac{P(z_t|\mathbf{Z}_{0:t-1},
M_j)P(M_j|\mathbf{Z}_{0:t-1})}{P(z_t|\mathbf{Z}_{0:t-1})} \nonumber \\
&\mu_j(t) = \frac{P(z_t|\mathbf{Z}_{0:t-1}, M_j)\mu_j(t-1)}{\sum_{j=1}^{r}
P(z_t|\mathbf{Z}_{0:t-1}, M_j)\mu_j(t-1) } \label{eq:choose_motion}
\end{align}
The probability of the current observation $z_t$ (the entire trajectory of
observation up to time $t$ is denoted by $\mathbf{Z}_{0:t-1}$) conditioned over a
specific articulated motion model and all the previous observations can be
represented by various methods. This probability for an Extended Kalman Filter
(EKF) based filtering algorithm is the probability of observation residual as
sampled from a normal distribution distributed with zero mean and innovation
covariance \cite{yaakov2001estimation}. One could pick a model by simply picking
the model with maximum probability. However, in a online estimation framework,
outliers might change the probability significantly in short term.
We pick a model based on probabilistic threshold, which is decided based on the
total number of the models $r$. As more and more observations are received, our 
estimation algorithm (Algorithm \ref{algo:estimate_motion_model}) chooses a specific 
model for each target object.

\begin{algorithm}
 \KwData{$\{M_j\}_{j=1}^r$, $z_t$, $\tau$}
 \KwResult{$\hat{M} \in \{M_j\}_{j=1}^r$, $C$, $P(q(t+1)|q(t))$ }
 initialization: $C_j = \{\}$, $M= \{\}$ \;
\While{$M= \{\}$} {
\ForAll{$M \in \{M_j\}_{j=1}^r$, }{
\eIf{$C_j$ is $\{ \}$ }{
Estimate $C_j$ \;
Estimate Temporal Structure \;
}{
Propagate state using EKF \;
Estimate $P(z_t|\mathbf{Z}_{0:t-1}, M_j)$ \;
}
}
\ForAll{$M \in \{M_j\}_{j=1}^r$, }{
Normalize to obtain $\mu_j(t)$ \;
  \If{$\mu_j(t)>\tau$}{
$\hat{M} = M_j$
   }
}
}
 \caption{Estimating the correct articulation model, associated configuration
 parameters, and motion variables}
\label{algo:estimate_motion_model}
\end{algorithm}
%Algorithm \ref{algo:estimate_motion_model} provides a pseudo-code for estimating
%a specific motion model. The algorithm requires definition of various
%articulated motion classes, a motion class threshold $\tau$ and observations. We decide on a specific motion model when the probability of that model as represented in Equation \ref{eq:choose_motion} is greater than the threshold $\tau$.

\section{SLAM for Dynamic Environment}
To demonstrate the need for articulation estimation with explicit time dependence, we propose an algorithm for performing SLAM in a dynamic scene.
\begin{figure}
\centering
\begin{tikzpicture}
\tikzstyle{main}=[circle, minimum size = 10mm, thick, draw =black!80, node distance = 10mm]
\tikzstyle{connect}=[-latex, thick]
\tikzstyle{box}=[rectangle, draw=black!100]
\coordinate (org_1) at (0,0);
  \node[main] (x_k_1)[right=of org_1] {$x_{t-1}$ };
  \node[main] (x_k) [right=of x_k_1] { $x_{t}$ };
  \node[main] (x_k_2) [right=of x_k] {$x_{t+1}$};
\coordinate[right=of x_k_2] (end_1);

  \node[main, fill = black!20] (u_k_1)[above=of x_k_1.west] {$u_{t-1}$ };
\node[main, fill = black!20] (u_k)[above=of x_k.west] {$u_{t}$ };
\node[main, fill = black!20] (u_k_2)[above=of x_k_2.west] {$u_{t+1}$ };

\node[main, fill = black!20] (z_k_1)[below=of x_k_1.east] {$z_{t-1}$ };
\node[main, fill = black!20] (z_k)[below=of x_k.east] {$z_{t}$ };
\node[main, fill = black!20] (z_k_2)[below=of x_k_2.east] {$z_{t+1}$ };

\node[main] (m_k_1)[below=of z_k_1.west] {$m_{t-1}$ };
\coordinate [left=of m_k_1](org_2);
\node[main] (m_k)[below=of z_k.west] {$m_{t}$ };
\node[main] (m_k_2)[below=of z_k_2.west] {$m_{t+1}$ };
\coordinate [right=of m_k_2](end_2);

\node[main] (v_k_1)[below=of m_k_1.east] {$v_{t-1}$ };
\node[main] (v_k)[below=of m_k.east] {$v_{t}$ };
\node[main] (v_k_2)[below=of m_k_2.east] {$v_{t+1}$ };

  \path 
 (org_1) edge [connect] (x_k_1)
(x_k_1) edge [connect] (x_k)
(x_k) edge [connect] (x_k_2)
(x_k_2) edge [connect] (end_1)

(u_k_1) edge [connect] (x_k_1)
(u_k) edge [connect] (x_k)
(u_k_2) edge [connect] (x_k_2)

(x_k_1) edge [connect] (z_k_1)
(x_k) edge [connect] (z_k)
(x_k_2) edge [connect] (z_k_2)

(m_k_1) edge [connect] (z_k_1)
(m_k) edge [connect] (z_k)
(m_k_2) edge [connect] (z_k_2)

 (org_2) edge [connect] (m_k_1)
(m_k_1) edge [connect] (m_k)
(m_k) edge [connect] (m_k_2)
(m_k_2) edge [connect] (end_2)

(v_k_1) edge [connect] (m_k_1)
(v_k) edge [connect] (m_k)
(v_k_2) edge [connect] (m_k_2);
\end{tikzpicture}
\caption{Graphical Model of the general SLAM problem. The known nodes are darker than the unknown nodes.}
\label{fig:graphical_model}
\end{figure}
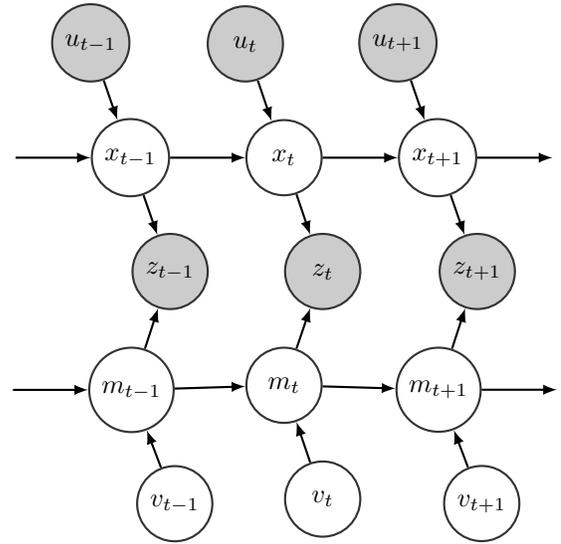
Figure \ref{fig:graphical_model} shows the graphical model of our general SLAM
problem where $x_t$, $u_t$, $z_t$, $m_t$, $v_t$ represent the robot state, input
to the robot, observation by the robot, state of the environment, and action of various
agents in the environment, respectively. Please note that $m$ in lower case is used to refer to the landmark while $M$ is used to refer to the type of motion model.

Basic SLAM algorithms assume the map $m_{t-1} \equiv m_t \equiv m$ to be static
and model the combination of robot state and map ($\{x_t, m\}$) as the state of the
estimation problem \cite{durrant2006simultaneous}. The associated estimation
problem only requires a motion model $P(x_t|x_{t-1},u_t)$ and an observation model
$P(z_t|x_t,m)$. The observation model assumes the observations to be conditionally independent given the map and the current vehicle state. The goal of the estimation process is to produce unbiased and consistent estimates (the expectation of mean squared errors should match the filter-calculated covariance) \cite{yaakov2001estimation}.

In contrast, for the dynamic SLAM problem, the state consists of a time-varying
map and the robot state. Hence,
the full estimation problem is:
\begin{align}
P(x_t,m_t|\mathbf{Z}_{0:t},\mathbf{U}_{0:t},\mathbf{V}_{0:t},x_0,m_0),
\end{align}
where $\mathbf{Z}_{0:t}$, $\mathbf{U}_{0:t}$ and $\mathbf{V}_{0:t}$ represent the set
of observations, robot control inputs and map control inputs from the start time
to time step $k$. It is assumed that the map is Markovian in nature which
implies that the start state of the map $m_0$ contains all the information
needed to make future prediction if actions of various agents in the world
$v_{t-1},...,v_{t+1}$ and their impact on the map are known.

\subsection{Motion Propagation} The motion propagation update models the evolution of state according
to the motion model. To write the equation concisely, let $A =\{
\mathbf{Z}_{0:t-1},\mathbf{U}_{0:t},\mathbf{V}_{0:t},x_0,m_0 \}$, then
\begin{align}
	P(x_t,m_t|A) &= 
\int \int P(x_t,x_{t-1},m_t,m_{t-1}|A) dx_{t-1} dm_{t-1} \nonumber \\
%&= \int \int P(x_k|x_{k-1},m_k,m_{k-1},A)P(x_{k-1},m_k,m_{k-1}|A) dx_{k-1}dm_{k-1} \nonumber \\
%&= \int \int P(x_k|x_{k-1},u_k)P(x_{k-1},m_k,m_{k-1}|A) dx_{k-1}dm_{k-1} \nonumber \\
%&= \int \int P(x_k|x_{k-1},u_k)P(m_k|x_{k-1},m_{k-1},A)P(x_{k-1},m_{k-1}|A)  dx_{k-1}dm_{k-1} \nonumber \\
&= \int \int P(x_t|x_{t-1},u_t)P(m_t|m_{t-1},v_{t-1})  \nonumber \\
& \qquad P(x_{t-1},m_{t-1}|A) dx_{t-1}dm_{t-1}  
\label{eq:time_update}
\end{align}

The independence relationship in the derivation of the time update in Eq.
\ref{eq:time_update} is due to the Bayesian network in Fig.
\ref{fig:graphical_model} in which each node is independent of its
non-descendants given the parents of that node. Given the structure of time
update, we need two motion models, one for the robot, $P(x_t|x_{t-1},u_t)$ and
another one for the world, $P(m_t|m_{t-1},v_{t-1})$. It can be clearly observed
that $P(m_t|m_{t-1},v_{t-1})$ for a static map is a Dirac Delta function and
integrates out in Eq. \ref{eq:time_update}. 

\subsection{Measurement Update} The measurement update uses Bayes formula to
update the state of the estimation problem given a new observation $z_t$ at time
step $t$. To write the equations concisely, let $B =\{
\mathbf{Z}_{0:t},\mathbf{U}_{0:t},\mathbf{V}_{0:t},x_0,m_0 \}$, then
 \begin{align}
P(x_t,m_t|B) &= \frac{P(z_t|x_t,m_t,A)P(x_t,m_t|A)}{P(z_t|A)} \nonumber\\
&=\frac{P(z_t|x_t,m_t)P(x_t,m_t|A)}{P(z_t|A)}
\label{eq:measurement_update}
\end{align}

Eq. \ref{eq:measurement_update}, together with Eq. \ref{eq:time_update}, defines
the complete recursive form of the SLAM algorithm for a dynamic environment. The
focus of current work is the representation of non-static environment using
articulation motion model in order to extend the standard SLAM algorithm with its static world assumption to a dynamic world.

\subsection{Dynamic World Representation} 
In this work, we demonstrate the SLAM algorithm for a feature based map. The
overall framework is extensible to other kinds of mapping algorithms such as,
dense maps. In feature based mapping, motion of each feature is assumed to be independent, given
the location of the feature at the previous time step. The state of the map,
$m_t$ is the collection of motion parameters of all the landmarks observed in
the scene. The true motion model for
each landmark in the scene is assumed to be one of the motion models $M \in
\{M_j\}^{r}_{j=1}$. 

\section{Articulated EKF SLAM}

\subsection{Robot Motion Model} We consider a robot with state $x_t =
(x,y,\theta)^T$ at time $t$ moving with constant linear velocity $v_t$ and
angular velocity $\omega_t$. The state of the robot at the next time step can be represented as 
\begin{align}
x_{t+1} = 
\begin{pmatrix}
x-\frac{v_t}{\omega_t} \sin{\theta} + \frac{v_t}{\omega_t}\sin(\theta+\omega_t \delta t) \\
y+\frac{v_t}{\omega_t} \cos{\theta} - \frac{v_t}{\omega_t}\cos(\theta+\omega_k \delta t) \\
\theta+\omega_k \delta t
\end{pmatrix}+ \mathcal{N}(0,N_t), \label{eq:robot_model}
\end{align},
where $\delta t$ is the width of the time step and $N_t$ is the error covariance
of the noise (zero mean Gaussian). Error covariance can be derived by
propagating the noise through the robot motion model and projecting the input
noise to the state space \cite{thrun2005probabilistic}.

If the angular velocity is close to zero, the robot model as represented in Eq. \ref{eq:robot_model} will be ill-conditioned. The model with zero angular velocity is given by 
\begin{align}
x_{t+1} = 
\begin{pmatrix}
x+v_t\delta t \cos(\theta) \\
y+v_t\delta t \sin(\theta)\\
\theta
\end{pmatrix}+ \mathcal{N}(0,N_t). \label{eq:robot_model}
\end{align}
Following the approximations proposed by Thrun et al.
\cite{thrun2005probabilistic}, the angular and linear velocities are generated
by a motion control unit $\hat{u}_t = (\hat{v}_t,\hat{\omega}_t)^T$ with zero mean additive Gaussian noise.
\begin{align}
\begin{pmatrix}
v_t\\
\omega_t
\end{pmatrix} & =
\begin{pmatrix}
\hat{v}_t\\
\hat{\omega}_t
\end{pmatrix}+
\mathcal{N}(0,S_t) \\
S_t & = 
\begin{pmatrix}
\alpha_1 \hat{v}_t^2 + \alpha_2 \hat{\omega}_t^2 &  0\\
0 & \alpha_3 \hat{v}_t^2 + \alpha_4 \hat{\omega}_t^2
\end{pmatrix}
\end{align}
where $\alpha_1,\alpha_2,\alpha_3,\alpha_4$ are the noise coefficients.

\subsection{Observation Model} The robot uses an ASUS Xtion camera which
provides depth estimates for all the features. Let the $j^{th}$ landmark be
located at position $m_j = (m_{j,x},m_{j,y},m_{j_z})^T$ within robot's sensor field of view.
Each observation can be written as
\begin{align}
z^i_t = 
R_{t}^T\begin{pmatrix}
	m_{j,x} - x\\
	m_{j,y} - y \\
	m_{j,z}
\end{pmatrix}+ \mathcal{N}(0,Q_k)
\end{align}
where $z^i_t$ is the $i^{th}$ observation of the $j^{th}$ landmark at time step
$t$ disturbed by zero mean Gaussian noise with covariance matrix $Q_k$ and
$R_{t}$ is the rotation matrix corresponding to rotation of angle $\theta_k$
around $z$ axis. It is assumed that the robot moves in the same plane and hence
its position in the $z$ axis is not a part of the state. However, our model is
readily extensible to general robot motion cases. 

\subsection{Jacobian Computation}
Extended Kalman Filtering (EKF) requires linearization of the robot's motion model to
ensure that the state propagation maintains Gaussianity of the state
distribution. In order to propagate the state, we estimate the Jacobian of the
state propagation model with respect to the state at time step $t$. The state Jacobian can be represented as
\begin{align}
J^{x_{t+1}}_{x_t} = 
\begin{pmatrix}
1 & 0 & -\frac{v_t}{\omega_t} \cos{\theta} +
	\frac{v_t}{\omega_t}\cos(\theta+\omega_t \delta t)\\
0 & 1 & -\frac{v_t}{\omega_t} \sin{\theta} +
	\frac{v_t}{\omega_t}\sin(\theta+\omega_t \delta t) \\
0 & 0 & 1
\end{pmatrix}
\end{align}
Furthermore, the error in the control space is projected onto the state
space, for which we compute the Jacobian of the state propagation model
with respect to the input $u_t$. 
%\begin{align}
%J_{u_k}^{x_{k+1}} = 
%\begin{pmatrix}
%\frac{-\sin{\theta}+\sin(\theta+\omega_k \delta t)}{\omega_k} & \frac{v_k(\sin{\theta}-\sin(\theta+\omega_k \delta t))}{\omega_k^2} + \frac{v_k\cos(\theta+\omega_k \delta t)}{\omega_k} \\
%\frac{\cos{\theta}-\cos(\theta+\omega_k \delta t)}{\omega_k} & \frac{-v_k(\cos{\theta}-\cos(\theta+\omega_k \delta t))}{\omega_k^2} + \frac{v_k\sin(\theta+\omega_k \delta t)}{\omega_k} \\
%0 & \delta t 
%\end{pmatrix}
%\end{align}

To assimilate each observation $z_i^t$, we compute the Jacobian of the
observation model with respect to the overall SLAM state, consisting of the
robot state as well as the motion parameters state associated with each
landmark. However, for the $i^{th}$ observation at time step $t$ of landmark $j$, the only relevant entries in the Jacobian matrix are the derivative of observation with respect to the robot states and the motion parameters state associated with landmark $j$. 
%The Jacobian of observation with respect to the robot state is 
%\begin{align}
%J^{z^i_k}_{x_k} = 
%\begin{pmatrix}
%	-\cos{\theta} & -\sin{\theta} &
%	-(m_{j,x}-x)\sin{\theta}+(m_{j,y}-y)\cos{\theta}\\
%	\sin{\theta} & -\cos{\theta} & 
%	-(m_{j,x}-x)\cos{\theta}-(m_{j,y}-y)\sin{\theta}\\
%	0 & 0 & 0
%\end{pmatrix}
%\end{align}
%and the Jacobian with respect to the motion parameters state is 
\begin{align}
	J^{z^i_t}_{q(t)} = 
\begin{pmatrix}
	\cos{\theta} & \sin{\theta} & 0\\
	-\sin{\theta} & \cos{\theta} & 0\\
	0 & 0 & 1
\end{pmatrix}J^{m_j}_{q(t)}
\end{align}
where $J^{m_j}_{q(t)}$ is the Jacobian of landmark observation with respect to
the motion parameters state $q(t)$ which is calculated based on the articulated
motion models as described in Section \ref{sec:jac_models}.

\begin{algorithm}
 \KwData{$\mu_{t-1}$, $\Sigma_{t-1}$, $u_t$, $\{M_j\}_{j=1}^r$, $z_t$, $\tau$}
 \KwResult{$\mu_t$, $\Sigma_t$}
Propagate Robot State and Covariance\;
Propagate Landmarks State and Covariance\;
\ForAll{$z^i_t \in z_t$}{
\uIf{$\hat{M} \neq \{ \}$ }{
Estimate Motion Model($\{M_j\}_{j=1}^r$, $z_t$, $\tau$)\;
}
\Else{Assimilate Observation\;}
}
\caption{Articulated EKF SLAM}\label{alg:articulated_slam}
\end{algorithm}
We summarize the entire articulated EKF SLAM approach to dynamic environments in
Algorithm \ref{alg:articulated_slam}.

\section{Experiments}
We analyze components of the proposed spatiotemporal articulation estimation
framework and also evaluate its predictive performance to explore dynamic
environments.
\subsection{Articulated Structure}\label{sec:configuration_estimation} 
\begin{figure}
\includegraphics[width=1\linewidth]{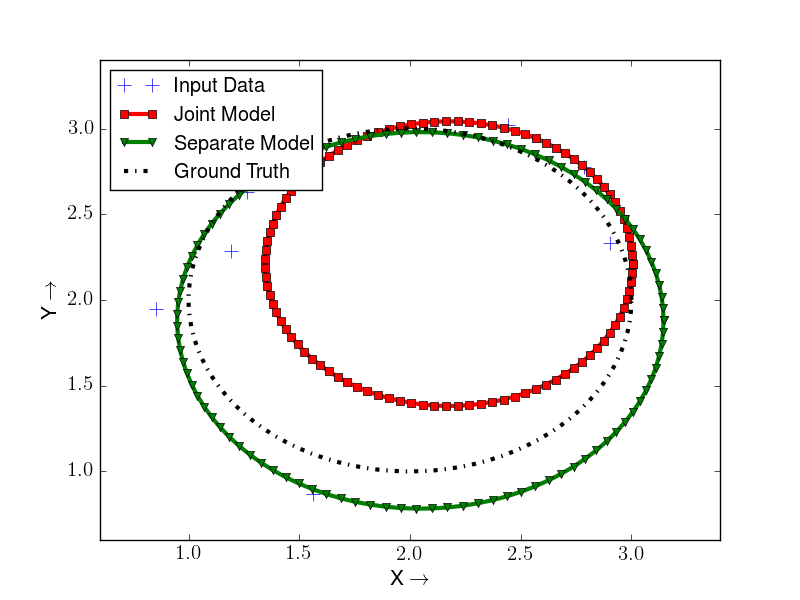}
\caption{Estimation of structure parameters for a 2D landmark undergoing revolute motion
centered at point $(2,2)$ with radius $1$. Zero mean Gaussian noise of $0.01$ variance is
assumed for both $X$ and $Y$ directions. }
\label{fig:joint_vs_separate_estimation}
\end{figure}

To elucidate
the effectiveness of separating motion parameters from structure parameters, 
we consider the structure estimation of a $2D$ landmark undergoing revolute motion.
In the current case, articulation structure refers to the center location and
radius of the landmark, while the motion parameter is $\theta$, the 
time-varying angle. For joint estimation, we choose a commonly used constant-velocity 
motion model,
\begin{align}
z_t = X_c+r[\cos(\theta_0+\delta T \omega),\sin(\theta_0+\delta T \omega)]^T \label{eq:2D_revolute}
\end{align}
where $z_t$ is the observed landmark position, $\theta_0$ is the initial angle,
$X_c$ is the landmark's center location and $\omega$ is the constant angular
velocity of the point. The estimation problem resulting from 
Eq. \ref{eq:2D_revolute} is non-linear in $\theta_0$ and $\omega$. On the other
hand, the proposed articulated structure estimation approach solves a well-conditioned 
problem of estimating a circle from points lying on a circle. 
We performed a Monte Carlo simulation and averaged errors across all the trials. 
Figure \ref{fig:joint_vs_separate_estimation} shows the estimation results using
least squares fitting between observed and predicted values of the landmark for
one specific run.
Table \ref{tab:error_joint_vs_separate} shows 
results from $500$ Monte Carlo runs of both methods. It can be 
observed that separating structure estimation from temporal modeling has 
significantly less error compared to joint estimation problem.

\begin{table}
\center
\begin{tabular}{| c | c| c| }
\hline
  Estimation & Center Error & Radius Error\\ \hline
  Joint & 0.71 & 0.09 \\ \hline
  Separate & 0.60 & 0.04 \\ \hline
\end{tabular}
\caption{$L2$ norm for center and radius error}
\label{tab:error_joint_vs_separate}
\end{table}

\subsection{Temporal Order} 
\begin{figure}
	\centering
	\includegraphics[width=0.8\linewidth]{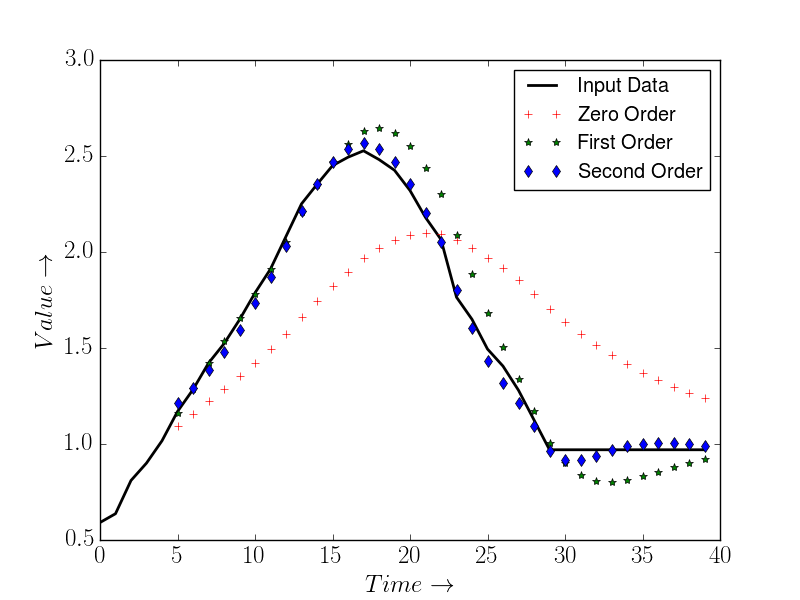}
\caption{Comparison of EKF filtering based state estimation for various orders of a motion parameter. }
\label{fig:order_model}
\end{figure}

After the evaluation of structure parameters, various 
orders of temporal models can be estimated using the approach outlined in
Section \ref{sec:temporal_structure}. To evaluate performance of various motion
continuity orders, we obtain raw angular trajectory of a
spring-loaded door and fit zeroth, first, and second order temporal models. We use the
temporal model in a EKF filtering framework with direct sensing of motion parameter
as observation model. This is equivalent
to observation of the landmark $z_t$ after the articulated structure is
estimated mapped via inverse kinematics. 
Figure \ref{fig:order_model} shows the motion parameter for different orders. 
It can be observed that increasing the order adds further flexibility. We choose
a first order continuity temporal model for rest of the experiments.

\subsection{Articulation Estimation}
\begin{figure}[t]
\begin{tikzpicture}[imgstyle/.style={draw,thick,anchor=south west,inner sep=1pt}]
 %\path[use as bounding box, draw] (0,-1) rectangle (\imgwidth, 0.34\imgwidth);
\node[imgstyle] (img1) at (0,0) {\includegraphics[width=0.32\linewidth]{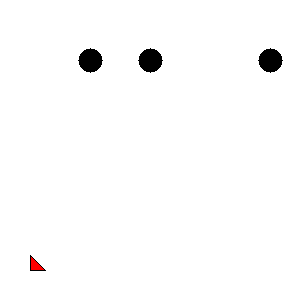}};
\node[imgstyle] (img2) at (0.333\linewidth,0) {\includegraphics[width=0.32\linewidth]
{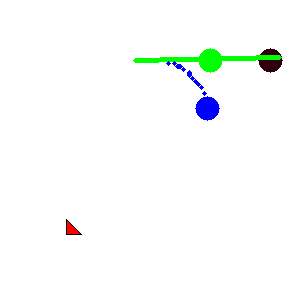}};
\node[imgstyle] (img3) at (0.666\linewidth,0) {\includegraphics[width=0.32\linewidth]{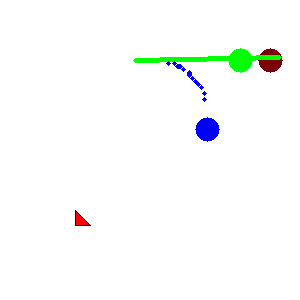}};
\begin{scope}[x={(img1.south east)}, y={(img1.north west)}]
\node (r) at (0.3, 0.1) {Robot};
\node (lmks) at (0.7, 0.6) {Landmarks};
\node (t1) at (0.8, 0.1) {t=0};
\end{scope}
\begin{scope}[x={(img2.south east)}, y={(img2.north west)}]
\node (t1) at (0.8, 0.1) {t=8};
\end{scope}
\begin{scope}[x={(img3.south east)}, y={(img3.north west)}]
\node (t1) at (0.8, 0.1) {t=10};
\end{scope}
\path ($(img1.south) + (-0.5, -0.5)$) node (ld) {Legend:};
\path 
(ld)
 ++ (1, 0) node [circle,fill=green] {} + (1.2, 0) node {Prismatic}
 ++ (3, 0) node [circle,fill=blue] {} + (1.2, 0) node {Revolute}
 ++ (3, 0) node [circle,fill=red] {} + (0.8, 0) node {Static}
;

 \end{tikzpicture}
    \caption{Frames at different time intervals of our simulation.
      Color of a landmark at a particular frame is the weighted sum of colors
      assigned to each motion model. The weights used are the probability of the
    landmark following that particular motion model as estimated by our algorithm. We also show the predicted trajectory of a landmark according to the estimated motion model.}
    \label{fig:results_articulation_estimation}
\end{figure}
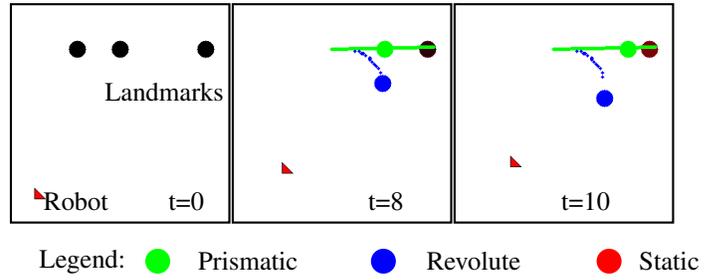

To test the articulation estimation framework, we simulated an environment with
one static, prismatic and revolute points each. We used a minimum of $7$ samples
to estimate configuration and initialize motion parameters. Figure
\ref{fig:results_articulation_estimation} shows the results for the articulation
estimation. Given sufficient observation, all the articulation models are
estimated correctly. However, static articulation takes the longest time to be
correctly estimated. This is because of the difficulty in separating static
landmark from a revolute and prismatic landmark with zero velocities.  

\subsection{Dynamic World SLAM} 
\begin{figure}
\subfloat[EKF
SLAM]{\includegraphics[width=0.49\linewidth,trim = 28mm 10mm 18mm
10mm,clip]{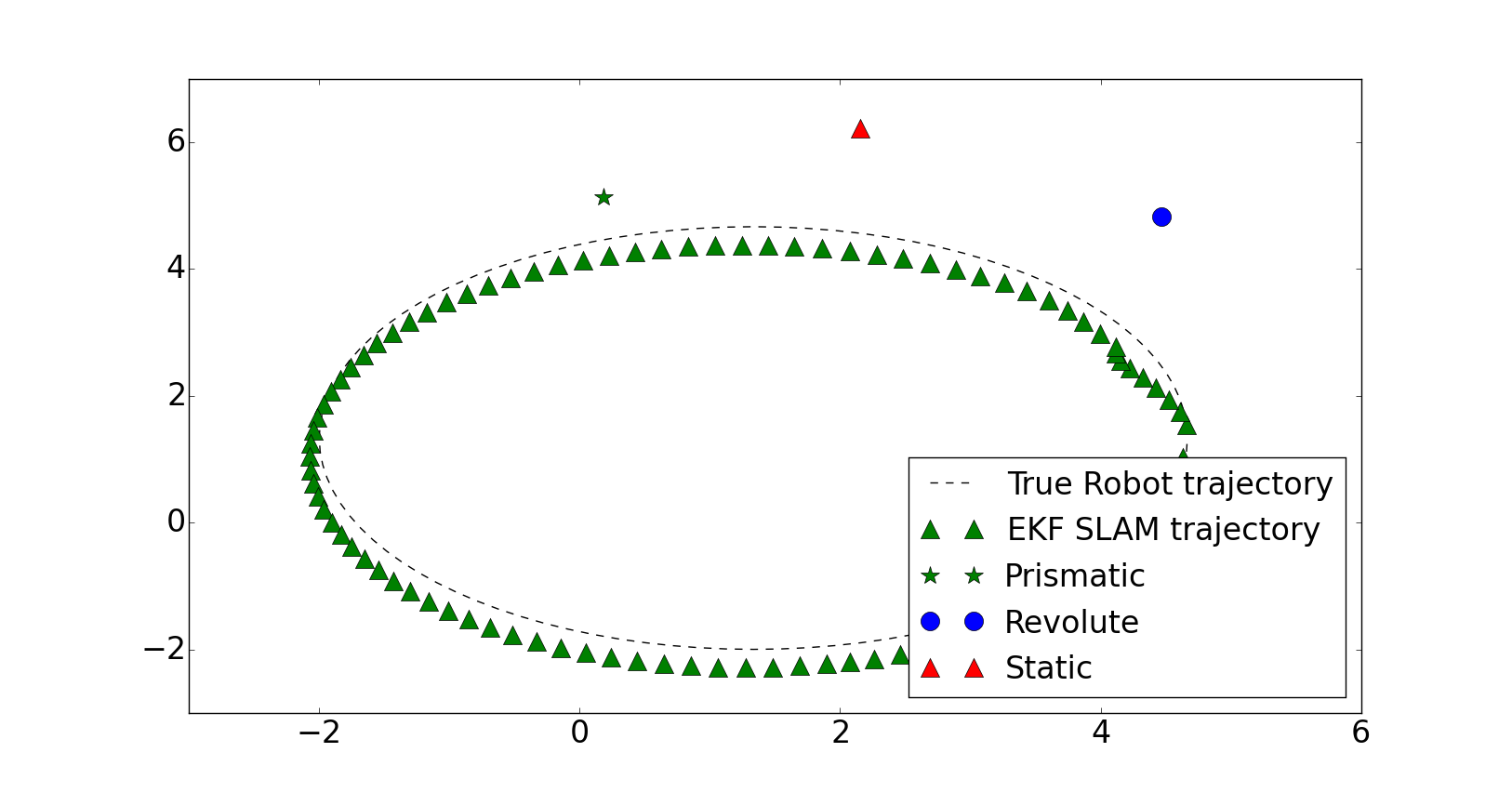}}
\subfloat[A-SLAM]{\includegraphics[width=0.49\linewidth,trim = 10mm 0mm 18mm
10mm,clip]{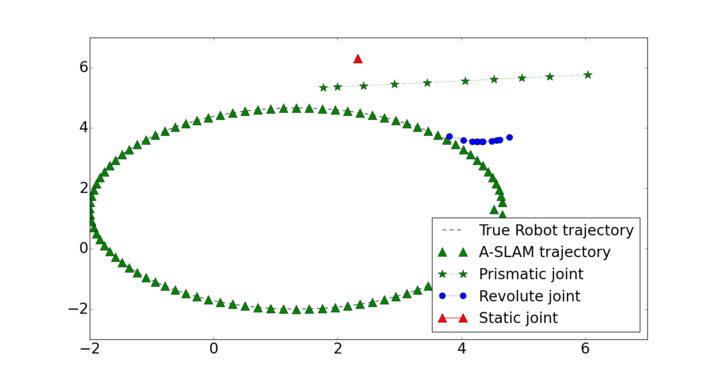}}
\caption{Demonstration of A-SLAM and SLAM algorithm at various time steps.}
\label{fig:articulated_ekf_model}
\end{figure}

In order to test our dynamic SLAM framework, 
we simulated a map with landmarks that are
either static, prismatic or revolute. A robot with limited field of view ($90$
degree cone of radius 4 meters) simulated readings from a depth sensor which were
then used to simultaneously localize the robot as well as to map the environment.
We used a total of 42 landmarks in the environment with $1$ revolute, $1$ prismatic 
and $40$ static landmarks. Both, the
revolute and prismatic landmarks were correctly identified with a threshold
$\tau = 0.6$. Figure \ref{fig:articulated_ekf_model} shows the robot
localization using both the standard EKF slam algorithm, a dynamic variant of standard EKF slam based on $x_t = x_{t-1} + (x_{t-1} - x_{t-2})$ and the articulated EKF
(A-SLAM) algorithm. 
\begin{figure*}
\subfloat[Static]{\includegraphics[width=0.32\linewidth,trim = 10mm 0mm 20mm 10mm,clip]{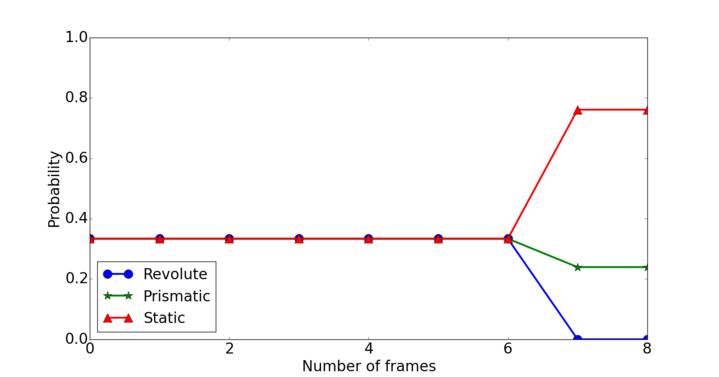}}
\subfloat[Prismatic]{\includegraphics[width=0.32\linewidth,trim = 10mm 0mm 20mm 10mm,clip]{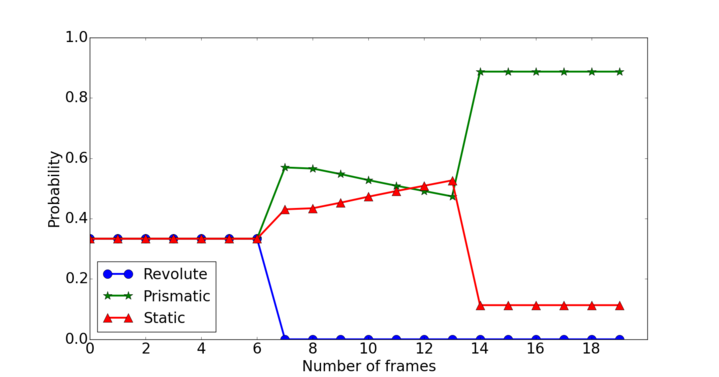}}
\subfloat[Revolute]{\includegraphics[width=0.32\linewidth,trim = 10mm 0mm 20mm 10mm,clip]{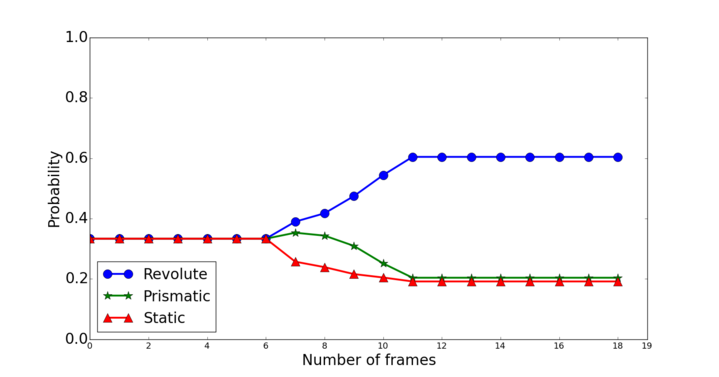}}
\caption{Model probabilities for articulated model selection for three different landmarks.}
\label{fig:articulated_ekf_model_map}
\end{figure*}

Table~\ref{tab:error_articulated_vs_normal} compares standard EKF SLAM algorithm and its dynamic variant
to the proposed A-SLAM algorithm. However, the standard EKF SLAM algorithm 
includes the landmark position in the state as opposed to 
to our algorithm where the SLAM state includes the motion parameters. 
As a result, we only compare the resulting localization estimates of the
robot using community accepted error metrics \cite{sturm2012benchmark}.
It is evident from the results that articulated structure estimation
significantly improves robot localization error.
Notably, we obtain such improvements in performance despite having
two non-static landmarks which violates the assumption of standard SLAM algorithms.

We also demonstrate our articulation model selection algorithm on the $3$
different landmarks as visualized in Figure \ref{fig:articulated_ekf_model}. 
The proposed algorithm 
needs a minimum of $7$ samples to fit articulated structure and
initialize temporal modeling. From the prismatic joint selection (Fig
\ref{fig:articulated_ekf_model_map} (b)), it can be observed that a maximum probability 
selection approach would fail at frame number $13$.

\begin{table}
\center
\begin{tabular}{| c | c| c|}
\hline
  Algorithm & Abs. Trajectory Error(ATE) & Relative Pose Error(RPE)\\ \hline
  EKF SLAM & 0.071 & 0.087 \\ \hline
  Dyn. SLAM & 0.049 & 0.051 \\ \hline
  A-SLAM & \textbf{0.014} & \textbf{0.003} \\ \hline
\end{tabular}
\caption{Comparison of localization error (observation
noise is sampled from a zero mean Gaussian $\mathcal{N}(0,0.04)$)) metrics for A-SLAM with EKF SLAM and its dynamic variant}
\label{tab:error_articulated_vs_normal}
\end{table}

\begin{figure}
	\includegraphics[width=1\linewidth]{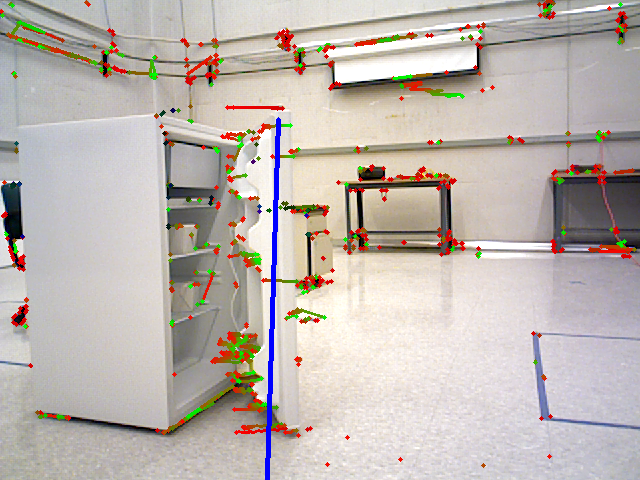}
	\caption{Tracked feature trajectories and overlaid revolute axis} 
	\label{fig:axis}
\end{figure}

\subsection{Indoor Objects} We acquired a RGBD video of an indoor scene with a
refrigerator door being articulated. We detect and track features by detecting
good features to track and match them in consecutive frames, using a pipeline
similar to the one proposed by Pillai et al. \cite{Pillai-RSS-14}. We use the
proposed articulation estimation framework and obtain the refrigerator axis by
clustering all the features that are classified as revolute. The visual result
is illustrated in Figure~\ref{fig:axis} which demonstrates correct prediction of
the articulated axis. Further, we present a comparison of localization estimates 
, similar to the previous experiment, in Table~\ref{tab:error_articulated_vs_normal_real}. 

\begin{table}
\center
\begin{tabular}{| c | c| c|}
\hline
  Algorithm & Abs. Trajectory Error(ATE) & Relative Pose Error(RPE)\\ \hline
  EKF SLAM & 0.059 & 0.073 \\ \hline
  Dyn. SLAM & 0.055 & 0.072 \\ \hline
  A-SLAM & 0.123 & 0.137 \\ \hline
\end{tabular}
\caption{Comparison of localization error 
metrics for A-SLAM with EKF SLAM and its dynamic variant on real world data.}
\label{tab:error_articulated_vs_normal_real}
\end{table}
\section{Conclusion} We presented a principled approach to articulated structure
estimation, which is essential for articulated motion representation in
unexplored environments. Our spatiotemporal articulation models can be estimated
by tracking a single landmark for 1-DOF joint. The proposed framework also
presents a passive approach to model temporal motion propagation. We
demonstrated that our approach outperforms the traditional SLAM algorithms by
integrating articulated structure estimation. Natural extensions of
the current framework will be to model rigid bodies and articulated chains.

\section*{Appendix}
\subsection{Articulated Motion Models} \label{sec:jac_models}
We briefly describe the articulated structure and motion parameters of the
articulated joints modeled in this paper.
\subsubsection{Revolute Joint} For a landmark moving according to revolute motion,
the articulation structure consists of the plane of motion $P$, location of center
and radius of the circle in that plane. The motion parameter to characterize the motion is simply the
angle $\phi_t$ of the current point with respect to the horizontal. The equation of
motion can be written as
\begin{align}
	m_j = (x_0+r\cos{\phi_t})v_1+(y_0+r\sin{\phi_t})v_2+P_0,
\end{align}
where $m_j$ is the current location of $j$ the landmark.
$v_1,v_2$ and $P_0$ are two perpendicular vectors in the plane and a point
on the plane $P$ respectively. $r$ is the radius of the circle and $\theta$ is
the angle with respect to horizontal. To estimate the articulated structure, we
first fit a plane in $3D$ and then fit a circle to the projection of points on
that plane using least squares. 
\subsubsection{Prismatic Joint}
Articulated structure estimation of a prismatic joint simply involves fitting a line to the
landmark positions over time. The equation of motion is given by
\begin{align}
	m_j = l_t\hat{n},
\end{align}
where $l_t$ is the time-varying motion parameter that measures length along a
line represented by $\hat{n}$.
\subsubsection{Static Joint}
For a static joint, we simply take the position of the static joint $P_0$ as the
structure and the motion variables models a perturbation about that
location. The equation of motion is given by
\begin{align}
	m_j = P_0+[1,1,1]^Tl_t,
\end{align}
where $l_t$ is the perturbation of the landmark at time step $t$ around the
static location.

%\subsection{Rotation Matrix of plane constrained motion}
%Rotation matrix in axis-angle representation can be written as
%\begin{align}
%	R_{t_0}^{t_1}& = I+\sin{\theta_t}K+(1-\cos{\theta_t})K^2
%\end{align}
%where $K$ is the cross-product matrix form of axes $\hat{w}$.
%Substituting the condition of translation to be perpendicular to plane normal in
%Equation \ref{eq:rotation_plane}, we get
%\begin{align}
%	x_0^T(R_{t_0}^{t_1}-I)^T\hat{n} & =0 \\ \nonumber
%	x_0^T(\sin{\theta_t}K+(1-\cos{\theta_t})K^2)^T\hat{n} & =0 \\ \nonumber
%	x_0^T(\sin{\theta_t}\hat{w}\times\hat{n}+(1-\cos{\theta_t})\hat{w}\times(\hat{w}\times\hat{n}))
%	& =0 \\ \nonumber
%	\sin{\theta_t}\hat{w}\times\hat{n}+(1-\cos{\theta_t})\hat{w}\times(\hat{w}\times\hat{n})
%	& =0 \label{eq:plane_mod}
%\end{align}
%Since $x_0^T \neq 0$ for all parameterization, Equation \ref{eq:plane_mod}
%implies $\hat{w}\times\hat{n} = 0$ because the combination of two linearly
%independent vectors can only be zero if their coefficients are zero. The
%coefficients $\sin{\theta_t}$ and $(1-\cos{theta_t})$ can not both be
%simultaneously zero except when $\theta_t = 2n\pi$. \nonumber

%\section*{Acknowledgments}

%% Use plainnat to work nicely with natbib. 

\bibliographystyle{plainnat}
\bibliography{articulated_slam_arxiv.bib}

\end{document}